\documentclass[11pt]{article}

\usepackage[final]{acl}

\usepackage{times}
\usepackage{latexsym}

\usepackage[T1]{fontenc}

\usepackage[utf8]{inputenc}

\usepackage{microtype}

\usepackage{inconsolata}

\usepackage{graphicx}
\usepackage{booktabs}
\usepackage{multirow}
\usepackage[table]{xcolor}
\usepackage{wrapfig}
\usepackage{amsmath}
\definecolor{resultred}{HTML}{F4CCCC}
\definecolor{resultgreen}{HTML}{D9EAD3}

\newlength{\originalintextsep}
\title{Upcycling Language Models for Multiple Specialized Domains
}

\author{
Mingzi Cao \quad Xi Wang \quad Nikolaos Aletras \\
School of Computer Science, University of Sheffield, United Kingdom \\
\texttt{\{mcao20,xi.wang,n.aletras\}@sheffield.ac.uk}
}

\begin{document}
\maketitle
\begin{abstract}

Effective multi-domain adaptation of large language models requires prioritizing relevant domain knowledge while minimizing interference from irrelevant domains.
Existing approaches commonly use Low-Rank Adaptation (LoRA) to train a separate adapter for each domain and combine these adapters within one model. However, they either weight adapters equally regardless of the input domain or dynamically weight adapters while still activating all of them including irrelevant ones. Additionally, while efficient, their low-rank structure limits performance.
In contrast, upcycling is a model expansion method that adds experts and a router to a dense Transformer layer offering larger capacity than LoRA approaches. 
In this paper, we propose Domain-Specialized Upcycling (\textsc{DS-Upcycling}) for multi-domain adaptation. We introduce a new domain-aware supervision that allows the router to effectively learn to distinguish domains by selecting the appropriate experts. During fine-tuning on each new domain, we update only the corresponding expert and router while freezing all other parameters.  
Experiments show that our method achieves the highest average performance across the adapted domains compared to baselines while retaining general performance. With just 256 examples per domain for router supervision, DS-Upcycling achieves 90--98\% routing accuracy after adaptation compared with 43--70\% for routers trained without our approach.

\end{abstract}

\begin{figure}[!t]
\begin{center}
\includegraphics[width=0.8\columnwidth]{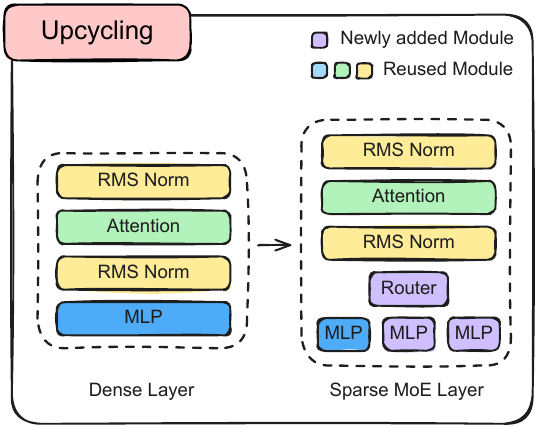}
\end{center}
  \caption{Illustration of Upcycling. Upcycling converts a pre-trained dense transformer layer into a sparse Mixture-of-Experts (MoE) layer.}
  \label{fig:main}
\end{figure}

\section{Introduction}

Large language models (LLMs) have demonstrated strong capabilities across a wide range of general tasks \citep{grattafiori2024llama3herdmodels, yang2025qwen3technicalreport}. However, domains involving consequential decisions such as biomedicine \citep{10.1093/bib/bbac409,singhal2023expertlevelmedicalquestionanswering} and law \citep{niklaus-etal-2024-multilegalpile,niklaus-etal-2025-lawinstruct} require specialized knowledge beyond general capabilities that is induced via fine-tuning on domain specific data. Multi-domain adaptation extends this specialization to multiple domains, aiming to integrate their specialized knowledge within a single model \citep{pmlr-v202-chen23aq,sukhbaatar2024branchtrainmix}.


Multi-domain adaptation methods suffer from catastrophic forgetting of general-domain capabilities \citep{li-etal-2024-revisiting,11151751} and cross-domain interference \citep{pmlr-v202-chen23aq,sukhbaatar2024branchtrainmix}.
Low-Rank Adaptation (LoRA) has been applied to address these challenges by freezing pre-trained weights and learning parameter updates through pairs of low-rank matrices \citep{hu2022lora}. Additionally, separate adapters are trained for each domain and then combined within one model for adaptation.
However, these approaches either weight adapters equally regardless of the input domain \citep{chitale2023task,wang-etal-2023-orthogonal,asadi2024doescombiningparameterefficientmodules} or dynamically weight adapters while still activating all of them including irrelevant ones \citep{dou-etal-2024-loramoe}. 
Beyond these limitations, LoRA’s adaptation performance is further constrained by its low-rank structure \citep{biderman2024lora,pletenev-etal-2025-much}.

Upcycling is a model expansion method that adds full multi-layer perceptron (MLP) experts and a router to a dense Transformer layer resulting in sparse Mixture-of-Experts (MoE) architecture (Figure~\ref{fig:main}). The new experts learn domain-specific knowledge without low-rank constraints. The router activates a subset of experts for each input and assigns weights to the selected experts. This offers a natural way to prioritize relevant domain knowledge and exclude irrelevant domain knowledge through selective expert activation \citep{switch_transformers,komatsuzaki2023sparse}. Despite these advantages, upcycling for multiple domain adaptation  remains largely underexplored. A few recent studies have explored multilingual adaptation \citep{10.1609/aaai.v39i24.34805,zhang-etal-2025-less}, but they rely on language-specific signals for router guidance that are not transferable to domains within a specific language (e.g., law or biomedical). 

In this work, we propose Domain-Specialized Upcycling (DS-Upcycling) for multi-domain adaptation (Figure~\ref{fig:ds_upcycling}). We first train the router with a domain-aware objective using only a small number of examples from each domain. This encourages general inputs to use the original MLP and discourages specialized inputs from using irrelevant experts. During adaptation to each  domain, we update only its corresponding expert and the router while freezing all other parameters. These domain-specific expert updates promote specialization by concentrating learning within the corresponding expert and limiting cross-domain interference. Our main contributions are as follows:

\begin{itemize}

\item We propose DS-Upcycling for multi-domain adaptation. It outperforms existing baselines across specialized and general domains in two different model families.

\item Our domain-aware router supervision uses just 256 examples per domain, requiring to only instantiating the routers.

\item Our ablation study validates the contribution of each DS-Upcycling component. DS-Upcycling achieves 90--98\% routing accuracy on specialized domains compared with 43--70\% for routers trained without it.

\end{itemize}

\section{Related Work}

\subsection{Multi-domain Adaptation}
LLM adaptation faces catastrophic forgetting of general-domain capabilities \citep{li-etal-2024-revisiting,11151751} and cross-domain interference \citep{pmlr-v202-chen23aq,sukhbaatar2024branchtrainmix}. To address these challenges, LoRA-based methods freeze pre-trained weights and learn updates through pairs of low-rank matrices \citep{hu2022lora}. A common strategy trains one adapter per domain and combines their outputs with equal weights \citep{chitale2023task,wang-etal-2023-orthogonal,asadi2024doescombiningparameterefficientmodules}. However, equal weighting prevents the model from prioritizing the adapter most relevant to the input domain. LoRAMoE addresses this limitation through a router that dynamically weights adapters \citep{dou-etal-2024-loramoe}. Yet it activates all adapters for every input and therefore cannot exclude irrelevant ones. Despite their efficiency, LoRA-based methods remain constrained by their low-rank architecture \citep{biderman2024lora,pletenev-etal-2025-much}. LLaMA-Pro instead inserts and trains additional transformer layers while freezing pre-trained parameters \citep{wu-etal-2024-llama}. Although it outperforms LoRA in specialized domains, its added layers are shared across domains and remain vulnerable to cross-domain interference. In contrast, our method trains full MLP experts for individual domains and selectively activates them.

\subsection{Model Upcycling}
\label{sec:related_upcycling}

\begin{figure*}[t]
\begin{center}
\includegraphics[width=0.8\textwidth]{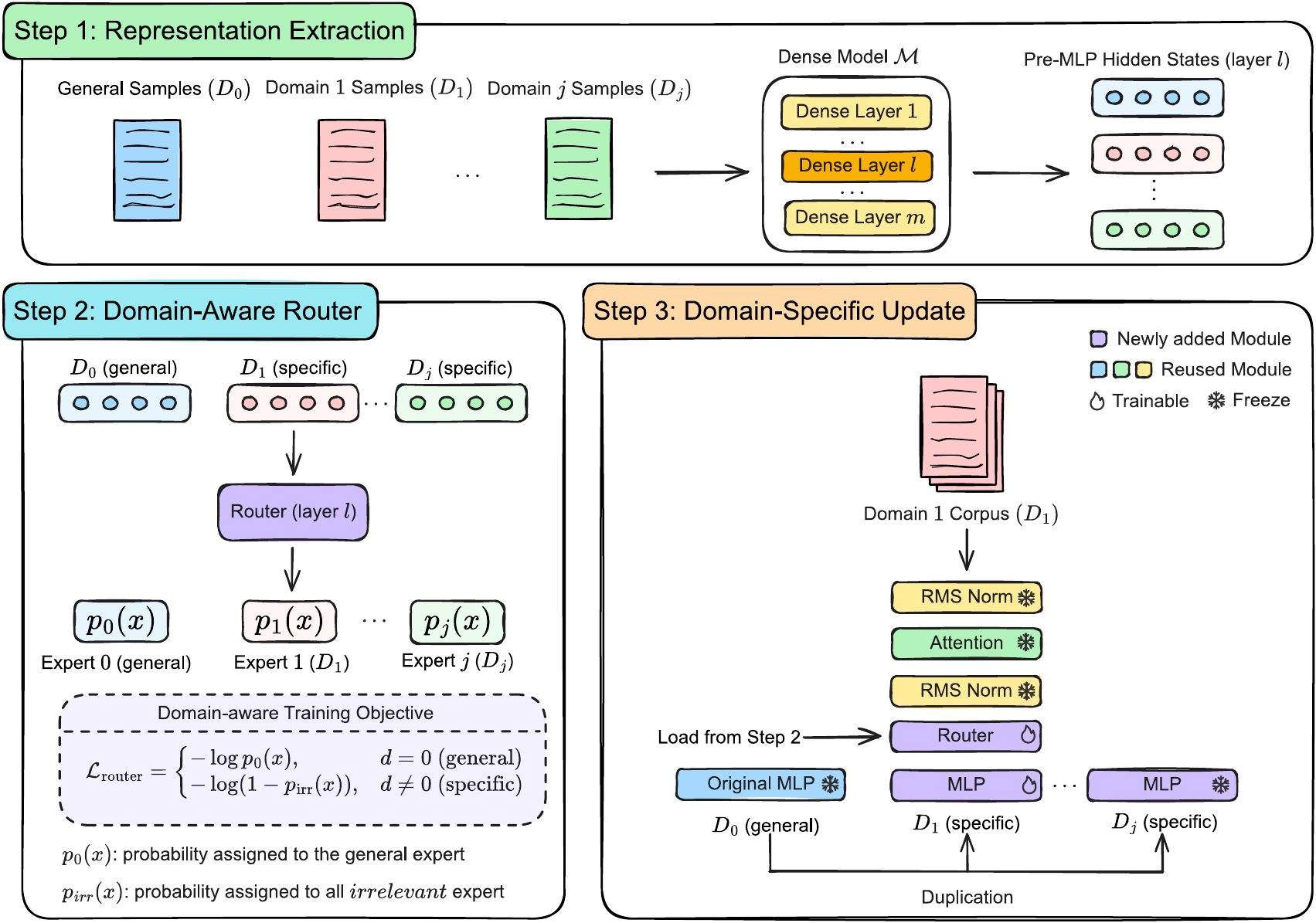}
\end{center}
  \caption{Illustration of DS-Upcycling. \textbf{Step 1:} Extract domain-specific hidden states. \textbf{Step 2:} Domain-aware router supervision favors the general expert for inputs from the general domain and suppresses irrelevant experts for inputs from specialized domains. \textbf{Step 3:} During adaptation to each specialized domain, update only its corresponding expert and the router while freezing all other parameters.}
  \label{fig:ds_upcycling}
\end{figure*}

\paragraph{Pre-training.}
Most upcycling work focuses on expert initialization for pre-training on large-scale general data. \citet{komatsuzaki2023sparse} find that duplicating the dense MLP performs best. However, identical initialization can hinder subsequent expert specialization \citep{wei2024skyworkmoedeepdivetraining,ICLR2025_d24b7366}. To address this issue, \citet{muennighoff2025olmoe} randomly replace some weights in each copied expert. \citet{ICLR2025_d24b7366} instead partially re-initialize experts using Gaussian distributions matched to pre-trained weight statistics. These methods use $>$200B tokens to train randomly initialized routers. However, they face two limitations in multi-domain adaptation. First, domain-specific datasets are much smaller than pre-training corpora and may be insufficient for randomly initialized routers to learn to distinguish domains from data alone. Second, making experts different at initialization does not ensure that they specialize in distinct domains unless subsequent training explicitly guides this specialization. Our method uses domain-aware router supervision to guide expert selection and domain-specific expert updates to promote specialization despite identical initialization.

\paragraph{Domain Adaptation.} 
Upcycling methods outperform other methods on multilingual adaptation \citep{10.1609/aaai.v39i24.34805}. \citet{10.1609/aaai.v39i24.34805} refine the router with language priors, and \citet{zhang-etal-2025-less} add a language classifier to guide routing, with the latter requiring architectural changes beyond a standard MoE layer. Therefore, both methods are not transferable to domain adaptation within the same language. They depend on the language of an input being identifiable, so that routing can be tied to language identity. Specialized domains such as legal and biomedical in English, share the same script and much of the same vocabulary and offer no comparable surface-level signals. In contrast, our method requires neither domain-specific priors nor architectural changes beyond a standard MoE layer and applies to arbitrary domains.

\section{Domain-Specialized Upcycling}

We propose Domain-Specialized Upcycling (DS-Upcycling), a new upcycling framework designed for multi-domain adaptation (Figure~\ref{fig:ds_upcycling}). It consists of three main steps: (1) domain-specific representation extraction, (2) domain-aware router supervision, and (3) domain-specific experts updates.

\subsection{Preliminaries}
\label{sec:preliminaries}

\paragraph{MoE.}
A sparse MoE layer contains $n$ MLP experts and a router with projection matrix $\mathbf{W}_{\text{router}}$. For input $x$, the router assigns expert weights:

{\small
\begin{equation*}
p_i(x)
=
\left[
\text{Softmax}\left(
\text{Top-}k(x^\top \mathbf{W}_{\text{router}})
\right)
\right]_i
\end{equation*}
}

\noindent Top-$k$ retains the $k<n$ largest logits and sets the rest to $-\infty$ \citep{switch_transformers}. The output is the weighted sum of selected experts.

{\small
\begin{equation*}
\text{MoE}(x)
=
\sum_{i=1}^{n} p_i(x)\text{MLP}_i(x)
\end{equation*}
}

\paragraph{Upcycling.}
Given a dense Transformer $\mathcal{M}$, upcycling retains the original MLP in each selected layer $l$ and adds $n-1$ new MLP experts and a router. All other modules remain unchanged:

{\small
\begin{equation*}
\text{MLP}^{(l)}(x)
\;\rightarrow\;
\text{MoE}^{(l)}(x)
\end{equation*}
}

\subsection{Representation Extraction}
\label{sec:step-1}

To extend upcycling to multi-domain adaptation, we extract hidden states from each domain for domain-aware router supervision before adaptation. Given $j$ specialized-domain datasets $\{D_1,\ldots,D_j\}$ and a general-domain dataset $D_0$, we sample a small set of examples from each and pass them through the dense model $\mathcal{M}$.

Let $L$ denote the set of layers to be upcycled. For each layer $l \in L$, let $\mathcal{X}_d^{(l)}$ denote the hidden states extracted immediately before the MLP from examples in $D_d$. We pair each hidden state with its domain label to construct the router training set:

{\small
\begin{equation*}
\mathcal{X}_{\mathrm{router}}^{(l)}
=
\left\{
(x,d) \mid
x \in \mathcal{X}_d^{(l)},\;
d \in \{0,\ldots,j\}
\right\}
\end{equation*}
}

\subsection{Domain-aware Router Supervision}
\label{sec:step-2}

Each upcycled layer contains the original dense MLP as its general expert and $j$ specialized experts corresponding to the target domains. For each layer $l \in L$, we train a separate router $\mathbf{W}^{(l)}_{\mathrm{router}}$ on $\mathcal{X}_{\mathrm{router}}^{(l)}$. This supervision requires only instantiating the routers, without involving the dense model. For a hidden state $x$ from domain $D_d$, we define the domain-aware objective as:

{\small
\begin{equation*}
\mathcal{L}_{\mathrm{router}}
=
\begin{cases}
-\log p_0(x), & d = 0,\\
-\log(1-p_{\mathrm{irr}}(x)), & d \in \{1,\ldots,j\}
\end{cases}
\end{equation*}
}

\noindent where $p_0(x)$ is the routing probability assigned to the general expert and $p_{\mathrm{irr}}(x)$ is the total routing probability assigned to specialized experts for domains other than $D_d$. For general-domain inputs, the objective encourages selection of the general expert. For specialized-domain inputs, it suppresses irrelevant experts while allowing both the corresponding specialized expert and the general expert to contribute.

\subsection{Expert Initialization}
\label{sec:step-3}

We initialize each specialized expert by copying the dense MLP following \citet{komatsuzaki2023sparse}. This is motivated by findings from prior work on expanding pre-trained models showing that initializing the expanded model to produce the same outputs as the original model benefits subsequent adaptation \citep{wu-etal-2024-llama,cao2025progressivedepthupscalingoptimal}. For each upcycled layer $l \in L$, the original MLP is copied to each specialized expert $\mathrm{MLP}^{(l)}_d
\leftarrow
\mathrm{MLP}^{(l)},
\forall\, l \in L,\ d \in \{1,\ldots,j\}.$


All experts initially produce identical outputs and the selected experts' routing weights sum to one. Their weighted output therefore matches the original MLP regardless of router selection.

\subsection{Domain-specific Expert Updates}
\label{sec:step-4}

DS-Upcycling has two training stages. First, each router is trained independently on the extracted hidden states using the domain-aware objective (\S\ref{sec:step-2}). Second, the upcycled model is adapted to each target domain. Inspired by prior work \citep{wang-etal-2024-expert,li-etal-2025-dynamic}, we update only the router and expert $d$ in each upcycled layer during adaptation to $D_d$ while freezing all other parameters. Restricting each expert's updates to its corresponding domain promotes expert specialization. This mitigates cross-domain interference while preserving the general expert's pre-trained knowledge.

\section{Experimental Setup}

\subsection{Base Models}

We use the pre-trained versions of \textit{Llama-3.1-8B} \citep{grattafiori2024llama3herdmodels} and \textit{Qwen-3-8B} \citep{yang2025qwen3technicalreport} as base models.

\subsection{Baselines}

We compare DS-Upcycling with a broad range of methods grouped into two categories: (1) dense baselines and (2) upcycling baselines.
Implementation and hyper-parameter details are provided in Appendices~\ref{App:Imp detail} and~\ref{App:hyp detail}.

\paragraph{Dense Baselines.}
These methods retain the dense Transformer layer architecture.
(1) \textit{Simple FT} updates all model parameters without architectural changes.
(2) \textit{LoRA} trains a separate adapter for each domain and combines their outputs with equal weights \citep{chitale2023task,wang-etal-2023-orthogonal,asadi2024doescombiningparameterefficientmodules}.
(3) \textit{LoRAMoE} uses a router to dynamically weight all adapter outputs \citep{dou-etal-2024-loramoe}.
(4) \textit{LLaMA-Pro} inserts new dense Transformer layers after each group of pre-trained layers and trains only the new layers \citep{wu-etal-2024-llama}.

\paragraph{Upcycling Baselines.}
These methods use upcycling to convert dense Transformer layers into sparse MoE layers.
They randomly initialize routers and update all parameters during adaptation but differ in expert initialization.
(1) \textit{Random Init} randomly initializes all weights of each new expert \citep{switch_transformers}.
(2) \textit{Duplication} initializes each new expert by copying the dense MLP \citep{komatsuzaki2023sparse}.
(3) \textit{Random Perturb} copies the dense MLP and randomly re-initializes a subset of its weights \citep{muennighoff2025olmoe}.
(4) \textit{Partial Init} copies the dense MLP and re-initializes a subset of its weights using Gaussians matched to their original distribution \citep{ICLR2025_d24b7366}.

\begin{table*}[!t]
\centering
\renewcommand{\arraystretch}{1.1}
\setlength{\tabcolsep}{1mm}
\setlength{\arrayrulewidth}{\lightrulewidth}
\arrayrulecolor{black}
\resizebox{0.98\textwidth}{!}{
\begin{tabular}{c|lccccccccc}
\toprule
\multicolumn{2}{c}{}
& \multicolumn{5}{c}{\textbf{Biomedical}}
& \multicolumn{4}{c}{\textbf{Legal}} \\
\cmidrule(l{2pt}r{2pt}){3-7}
\cmidrule(l{2pt}r{2pt}){8-11}
\multicolumn{2}{c}{\textbf{Methods}}
& \textbf{MedMCQA}
& \textbf{MedQA}
& \textbf{Bio-MMLU}
& \textbf{PubMedQA}
& \textbf{Average}
& \textbf{Law-MMLU}
& \textbf{CaseHOLD}
& \textbf{ECtHR}
& \textbf{Average} \\
\midrule

\multicolumn{1}{c}{} & \textit{Llama-3.1-8B}
& $56.34_{0.19}$
& $\underline{60.36_{0.12}}$
& $\underline{72.37_{0.28}}$
& $75.67_{0.12}$
& $66.18_{0.05}$
& $\underline{67.98_{0.27}}$
& $52.31_{0.22}$
& $60.60_{0.14}$
& $60.30_{0.07}$ \\
\midrule

\multirow{4}{*}{\rotatebox[origin=c]{90}{\textbf{Dense}}}
& Simple FT
& \cellcolor{resultred}$40.11_{0.10}$
& \cellcolor{resultred}$43.23_{0.30}$
& \cellcolor{resultred}$48.75_{0.12}$
& \cellcolor{resultred}$70.67_{0.12}$
& \cellcolor{resultred}$50.69_{0.05}$
& \cellcolor{resultred}$45.98_{3.90}$
& \cellcolor{resultgreen}$76.14_{0.05}$
& \cellcolor{resultred}$39.52_{0.07}$
& \cellcolor{resultred}$53.88_{1.31}$ \\

& LoRA
& \cellcolor{resultred}$56.04_{0.05}$
& \cellcolor{resultred}$56.64_{0.34}$
& \cellcolor{resultred}$66.98_{0.10}$
& \cellcolor{resultgreen}$\underline{78.33_{0.12}}$
& \cellcolor{resultred}$64.50_{0.07}$
& \cellcolor{resultred}$65.33_{0.05}$
& \cellcolor{resultgreen}$70.39_{0.14}$
& \cellcolor{resultgreen}$\underline{62.71_{0.14}}$
& \cellcolor{resultgreen}$\underline{66.14_{0.02}}$ \\

& LoRAMoE
& \cellcolor{resultgreen}$\mathbf{60.20_{0.11}}$
& \cellcolor{resultred}$57.27_{0.16}$
& \cellcolor{resultred}$71.10_{0.08}$
& \cellcolor{resultgreen}$\mathbf{78.53_{0.12}}$
& \cellcolor{resultgreen}$\underline{66.77_{0.02}}$
& \cellcolor{resultred}$67.49_{0.02}$
& \cellcolor{resultgreen}$71.65_{0.07}$
& \cellcolor{resultred}$58.95_{0.07}$
& \cellcolor{resultgreen}$66.03_{0.04}$ \\

& LLaMA-Pro
& \cellcolor{resultgreen}$57.34_{0.03}$
& \cellcolor{resultred}$58.29_{0.27}$
& \cellcolor{resultred}$67.49_{0.45}$
& \cellcolor{resultgreen}$78.07_{0.12}$
& \cellcolor{resultred}$65.30_{0.15}$
& \cellcolor{resultred}$63.92_{0.09}$
& \cellcolor{resultgreen}$53.76_{0.08}$
& \cellcolor{resultgreen}$\mathbf{62.91_{0.07}}$
& \cellcolor{resultred}$60.20_{0.03}$ \\
\midrule

\multirow{5}{*}{\rotatebox[origin=c]{90}{\textbf{Upcycling}}}
& Random Init.
& \cellcolor{resultred}$40.74_{0.10}$
& \cellcolor{resultred}$41.29_{0.23}$
& \cellcolor{resultred}$49.68_{0.16}$
& \cellcolor{resultred}$69.93_{0.23}$
& \cellcolor{resultred}$50.41_{0.02}$
& \cellcolor{resultred}$44.39_{5.57}$
& \cellcolor{resultgreen}$76.41_{0.02}$
& \cellcolor{resultred}$49.05_{0.26}$
& \cellcolor{resultred}$56.62_{1.87}$ \\

& Duplication
& \cellcolor{resultred}$40.25_{0.05}$
& \cellcolor{resultred}$43.05_{0.34}$
& \cellcolor{resultred}$47.90_{0.25}$
& \cellcolor{resultred}$72.67_{0.31}$
& \cellcolor{resultred}$50.97_{0.02}$
& \cellcolor{resultred}$46.24_{4.12}$
& \cellcolor{resultgreen}$\underline{76.67_{0.03}}$
& \cellcolor{resultred}$44.27_{0.29}$
& \cellcolor{resultred}$55.73_{1.42}$ \\

& Random Perturb.
& \cellcolor{resultred}$37.60_{0.14}$
& \cellcolor{resultred}$40.46_{0.08}$
& \cellcolor{resultred}$46.28_{0.11}$
& \cellcolor{resultred}$73.40_{0.00}$
& \cellcolor{resultred}$49.43_{0.07}$
& \cellcolor{resultred}$41.96_{6.43}$
& \cellcolor{resultgreen}$\mathbf{76.85_{0.04}}$
& \cellcolor{resultred}$40.35_{0.21}$
& \cellcolor{resultred}$53.05_{2.12}$ \\

& Partial Init.
& \cellcolor{resultred}$39.24_{0.14}$
& \cellcolor{resultred}$39.43_{0.28}$
& \cellcolor{resultred}$47.14_{0.07}$
& \cellcolor{resultred}$74.47_{0.12}$
& \cellcolor{resultred}$50.07_{0.06}$
& \cellcolor{resultred}$45.10_{4.70}$
& \cellcolor{resultgreen}$76.52_{0.08}$
& \cellcolor{resultred}$43.77_{0.14}$
& \cellcolor{resultred}$55.13_{1.64}$ \\

& \textbf{DS-Upcycling (Ours)}
& \cellcolor{resultgreen}$\underline{59.76_{0.11}}$
& \cellcolor{resultgreen}$\mathbf{61.51_{0.34}}^*$
& \cellcolor{resultgreen}$\mathbf{75.08_{0.05}}^*$
& \cellcolor{resultgreen}$77.33_{0.12}$
& \cellcolor{resultgreen}$\mathbf{68.42_{0.08}}^*$
& \cellcolor{resultgreen}$\mathbf{68.79_{0.31}}$
& \cellcolor{resultgreen}$75.52_{0.10}$
& \cellcolor{resultgreen}$62.58_{0.26}$
& \cellcolor{resultgreen}$\mathbf{68.96_{0.14}}^*$ \\

\midrule
\midrule

\multicolumn{1}{c}{} & \textit{Qwen-3-8B}
& $62.49_{0.08}$
& $\underline{65.54_{0.09}}$
& $80.11_{0.17}$
& $76.53_{0.23}$
& $71.17_{0.09}$
& $\underline{74.77_{0.16}}$
& $63.59_{0.08}$
& $60.81_{0.36}$
& $66.39_{0.20}$ \\
\midrule

\multirow{4}{*}{\rotatebox[origin=c]{90}{\textbf{Dense}}}
& Simple FT
& \cellcolor{resultred}$60.42_{0.11}$
& \cellcolor{resultred}$61.46_{0.23}$
& \cellcolor{resultred}$78.00_{0.28}$
& \cellcolor{resultgreen}$78.33_{0.23}$
& \cellcolor{resultred}$69.55_{0.06}$
& \cellcolor{resultred}$62.13_{0.36}$
& \cellcolor{resultgreen}$\underline{83.31_{0.02}}$
& \cellcolor{resultgreen}$61.88_{0.00}$
& \cellcolor{resultgreen}$69.11_{0.13}$ \\

& LoRA
& \cellcolor{resultred}$59.73_{0.14}$
& \cellcolor{resultred}$50.64_{0.23}$
& \cellcolor{resultred}$78.71_{0.14}$
& \cellcolor{resultgreen}$\underline{79.27_{0.12}}$
& \cellcolor{resultred}$67.09_{0.15}$
& \cellcolor{resultred}$70.50_{0.42}$
& \cellcolor{resultgreen}$72.44_{0.11}$
& \cellcolor{resultgreen}$\underline{64.52_{0.07}}$
& \cellcolor{resultgreen}$69.15_{0.13}$ \\

& LoRAMoE
& \cellcolor{resultred}$61.74_{0.07}$
& \cellcolor{resultred}$62.32_{0.05}$
& \cellcolor{resultgreen}$\mathbf{82.74_{0.27}}$
& \cellcolor{resultgreen}$\mathbf{79.60_{0.20}}$
& \cellcolor{resultgreen}$\underline{71.60_{0.09}}$
& \cellcolor{resultred}$71.79_{0.19}$
& \cellcolor{resultgreen}$75.60_{0.06}$
& \cellcolor{resultgreen}$63.70_{0.26}$
& \cellcolor{resultgreen}$70.36_{0.13}$ \\

& LLaMA-Pro
& \cellcolor{resultgreen}$\underline{63.42_{0.06}}$
& \cellcolor{resultred}$64.44_{0.36}$
& \cellcolor{resultred}$78.39_{0.17}$
& \cellcolor{resultred}$75.40_{0.00}$
& \cellcolor{resultred}$70.41_{0.15}$
& \cellcolor{resultred}$71.34_{0.13}$
& \cellcolor{resultgreen}$74.31_{0.05}$
& \cellcolor{resultgreen}$63.82_{0.14}$
& \cellcolor{resultgreen}$69.82_{0.01}$ \\
\midrule

\multirow{5}{*}{\rotatebox[origin=c]{90}{\textbf{Upcycling}}}
& Random Init.
& \cellcolor{resultred}$59.38_{0.09}$
& \cellcolor{resultred}$61.12_{0.21}$
& \cellcolor{resultred}$78.29_{0.25}$
& \cellcolor{resultgreen}$76.80_{0.00}$
& \cellcolor{resultred}$68.90_{0.12}$
& \cellcolor{resultred}$66.22_{0.24}$
& \cellcolor{resultgreen}$82.51_{0.09}$
& \cellcolor{resultgreen}$61.84_{0.07}$
& \cellcolor{resultgreen}$70.19_{0.12}$ \\

& Duplication
& \cellcolor{resultred}$60.27_{0.11}$
& \cellcolor{resultred}$63.34_{0.12}$
& \cellcolor{resultred}$77.45_{0.07}$
& \cellcolor{resultgreen}$78.20_{0.53}$
& \cellcolor{resultred}$69.81_{0.15}$
& \cellcolor{resultred}$62.34_{0.47}$
& \cellcolor{resultgreen}$82.80_{0.09}$
& \cellcolor{resultred}$60.52_{0.12}$
& \cellcolor{resultgreen}$68.55_{0.14}$ \\

& Random Perturb.
& \cellcolor{resultred}$60.52_{0.03}$
& \cellcolor{resultred}$62.58_{0.12}$
& \cellcolor{resultred}$78.05_{0.37}$
& \cellcolor{resultred}$76.47_{0.12}$
& \cellcolor{resultred}$69.40_{0.13}$
& \cellcolor{resultred}$64.20_{0.09}$
& \cellcolor{resultgreen}$81.39_{0.05}$
& \cellcolor{resultgreen}$63.74_{0.00}$
& \cellcolor{resultgreen}$69.77_{0.01}$ \\

& Partial Init.
& \cellcolor{resultred}$59.57_{0.03}$
& \cellcolor{resultred}$62.79_{0.50}$
& \cellcolor{resultred}$79.65_{0.34}$
& \cellcolor{resultgreen}$77.53_{0.12}$
& \cellcolor{resultred}$69.89_{0.19}$
& \cellcolor{resultred}$63.86_{0.42}$
& \cellcolor{resultgreen}$\mathbf{83.49_{0.11}}$
& \cellcolor{resultgreen}$\mathbf{66.38_{0.07}}$
& \cellcolor{resultgreen}$\underline{71.24_{0.14}}$ \\

& \textbf{DS-Upcycling (Ours)}
& \cellcolor{resultgreen}$\mathbf{66.62_{0.05}}^*$
& \cellcolor{resultgreen}$\mathbf{69.05_{0.27}}^*$
& \cellcolor{resultgreen}$\underline{80.41_{0.22}}$
& \cellcolor{resultgreen}$77.27_{0.31}$
& \cellcolor{resultgreen}$\mathbf{73.34_{0.09}}^*$
& \cellcolor{resultgreen}$\mathbf{75.76_{0.16}}^*$
& \cellcolor{resultgreen}$78.12_{0.11}$
& \cellcolor{resultgreen}$62.58_{0.14}$
& \cellcolor{resultgreen}$\mathbf{72.15_{0.09}}^*$ \\

\bottomrule
\end{tabular}
}
\caption{Performance on specialized domains. \textbf{Bold}/\underline{underlined} denote best/second-best results per task. \colorbox[HTML]{D9EAD3}{Green}/\colorbox[HTML]{F4CCCC}{Red} indicates scores above/below the untrained base model. When our method ranks first, $^*$ marks significant improvements over the second-best method (paired $t$-test, $p<0.05$).}
\label{tab:8B-Specialized}
\end{table*}

\subsection{Training Data}
\label{sec:data sampling}
We evaluate DS-Upcycling and all baselines in legal and biomedical domains as they are widely explored in prior NLP research \citep{10.1145/3458754,10.1093/bib/bbac409,singhal2023expertlevelmedicalquestionanswering,chalkidis-etal-2019-neural,10.1145/3462757.3466088,NEURIPS2022_bc218a0c,niklaus-etal-2024-multilegalpile}. Both domains require specialized knowledge beyond general capabilities.
We sample uniformly across task subsets within each dataset to obtain balanced training sets of 393K examples per domain.

\paragraph{Legal.} We use the English subset of LawInstruct that covers diverse legal tasks \citep{niklaus-etal-2025-lawinstruct}.

\paragraph{Biomedical.} We train on MedINST \citep{han-etal-2024-medinst} using the task subsets included in MultiMedQA \citep{singhal2022largelanguagemodelsencode}. \\

\noindent For domain-aware router supervision, we extract hidden states from 256 randomly selected examples per domain (\S\ref{sec:step-1}). Legal and biomedical examples come from their respective training  datasets. General-domain examples come from Alpaca-GPT4 \citep{peng2023instructiontuninggpt4}.

\subsection{Evaluation Tasks}

We use the legal, biomedical and general-domain benchmarks from \citet{williams-etal-2026-compressing}. The general benchmarks assess catastrophic forgetting. 
For all datasets, we report the mean and standard deviation across three random seeds.

\paragraph{Biomedical.}
We use PubMedQA \citep{jin-etal-2019-pubmedqa}, MedQA \citep{app11146421}, MedMCQA \citep{pmlr-v174-pal22a} tasks, and relevant subsets from MMLU \citep{hendrycks2021measuring} such as anatomy, clinical knowledge, college medicine, medical genetics, professional medicine, and college biology.

\paragraph{Legal.}
We use CaseHOLD \citep{10.1145/3462757.3466088} and ECtHR (Task A) \citep{chalkidis-etal-2019-neural} from LexGLUE \citep{chalkidis-etal-2022-lexglue} and Legal-MMLU, covering jurisprudence, professional law, and international law \citep{hendrycks2021measuring}.

\paragraph{General.} We use ARC \citep{clark2018thinksolvedquestionanswering}, BoolQ \citep{clark-etal-2019-boolq}, HellaSwag \citep{zellers-etal-2019-hellaswag}, LAMBADA \citep{paperno-etal-2016-lambada}, OpenBookQA \citep{banerjee-etal-2019-careful}, PIQA \citep{Bisk2020}, RTE \citep{10.1007/11736790_9}, StoryCloze \citep{mostafazadeh-etal-2016-corpus} and WinoGrande \citep{10.1145/3474381}.

\section{Results}
\label{sec:results}

\begin{table*}[!h]
\centering
\renewcommand{\arraystretch}{1.4}
\setlength{\tabcolsep}{1mm}
\setlength{\arrayrulewidth}{\lightrulewidth}
\arrayrulecolor{black}
\resizebox{0.98\textwidth}{!}{
\begin{tabular}{c|lccccccccccc}
\toprule
\multicolumn{2}{c}{\textbf{Methods}}
& \textbf{ARC-C}
& \textbf{ARC-E}
& \textbf{BoolQ}
& \textbf{HellaSwag}
& \textbf{LAMBADA}
& \textbf{OpenBookQA}
& \textbf{PIQA}
& \textbf{RTE}
& \textbf{WinoGrande}
& \textbf{StoryCloze}
& \textbf{Average} \\
\midrule

\multicolumn{1}{c}{}
& \textit{Llama-3.1-8B}
& $51.37_{0.00}$
& $81.50_{0.14}$
& $82.08_{0.05}$
& $\mathbf{60.02_{0.04}}$
& $67.24_{0.03}$
& $33.20_{0.00}$
& $\mathbf{80.09_{0.11}}$
& $\underline{70.52_{0.42}}$
& $\underline{73.72_{0.28}}$
& $78.14_{0.10}$
& $67.79_{0.08}$ \\
\midrule

\multirow{4}{*}{\rotatebox[origin=c]{90}{\textbf{Dense}}}
& Simple FT
& \cellcolor{resultred}$32.91_{0.26}$
& \cellcolor{resultred}$61.50_{0.10}$
& \cellcolor{resultred}$68.63_{0.26}$
& \cellcolor{resultred}$42.55_{0.02}$
& \cellcolor{resultred}$34.17_{0.11}$
& \cellcolor{resultred}$24.20_{0.00}$
& \cellcolor{resultred}$69.06_{0.25}$
& \cellcolor{resultred}$56.20_{0.21}$
& \cellcolor{resultred}$61.14_{0.23}$
& \cellcolor{resultred}$65.65_{0.11}$
& \cellcolor{resultred}$51.60_{0.04}$ \\

& LoRA
& \cellcolor{resultgreen}$52.10_{0.21}$
& \cellcolor{resultred}$80.64_{0.11}$
& \cellcolor{resultred}$80.37_{0.08}$
& \cellcolor{resultred}$57.40_{0.04}$
& \cellcolor{resultred}$65.08_{0.06}$
& \cellcolor{resultgreen}$\mathbf{37.60_{0.20}}$
& \cellcolor{resultred}$77.77_{0.08}$
& \cellcolor{resultred}$61.13_{0.21}$
& \cellcolor{resultred}$71.95_{0.09}$
& \cellcolor{resultgreen}$\mathbf{79.24_{0.08}}$
& \cellcolor{resultred}$66.33_{0.00}$ \\

& LoRAMoE
& \cellcolor{resultgreen}$\mathbf{53.73_{0.18}}$
& \cellcolor{resultgreen}$\underline{82.44_{0.13}}$
& \cellcolor{resultgreen}$\underline{82.68_{0.12}}$
& \cellcolor{resultred}$58.84_{0.04}$
& \cellcolor{resultgreen}$\mathbf{71.32_{0.04}}$
& \cellcolor{resultgreen}$\underline{35.73_{0.12}}$
& \cellcolor{resultred}$77.78_{0.03}$
& \cellcolor{resultred}$65.82_{0.21}$
& \cellcolor{resultred}$73.59_{0.12}$
& \cellcolor{resultred}$77.28_{0.08}$
& \cellcolor{resultgreen}$\underline{67.92_{0.03}}$ \\

& LLaMA-Pro
& \cellcolor{resultgreen}$\underline{53.19_{0.20}}$
& \cellcolor{resultred}$79.69_{0.14}$
& \cellcolor{resultred}$81.83_{0.05}$
& \cellcolor{resultred}$55.74_{0.01}$
& \cellcolor{resultgreen}$\underline{70.88_{0.10}}$
& \cellcolor{resultgreen}$35.67_{0.12}$
& \cellcolor{resultred}$78.62_{0.00}$
& \cellcolor{resultred}$66.43_{0.36}$
& \cellcolor{resultred}$71.06_{0.12}$
& \cellcolor{resultred}$76.09_{0.17}$
& \cellcolor{resultred}$66.92_{0.06}$ \\
\midrule

\multirow{5}{*}{\rotatebox[origin=c]{90}{\textbf{Upcycling}}}
& Random Init.
& \cellcolor{resultred}$33.79_{0.15}$
& \cellcolor{resultred}$64.72_{0.11}$
& \cellcolor{resultred}$70.37_{0.03}$
& \cellcolor{resultred}$41.84_{0.01}$
& \cellcolor{resultred}$31.74_{0.07}$
& \cellcolor{resultred}$23.53_{0.12}$
& \cellcolor{resultred}$69.91_{0.22}$
& \cellcolor{resultred}$54.99_{0.42}$
& \cellcolor{resultred}$60.43_{0.09}$
& \cellcolor{resultred}$67.50_{0.00}$
& \cellcolor{resultred}$51.88_{0.07}$ \\

& Duplication
& \cellcolor{resultred}$35.55_{0.10}$
& \cellcolor{resultred}$61.95_{0.00}$
& \cellcolor{resultred}$74.03_{0.11}$
& \cellcolor{resultred}$42.75_{0.06}$
& \cellcolor{resultred}$38.55_{0.07}$
& \cellcolor{resultred}$23.73_{0.12}$
& \cellcolor{resultred}$68.99_{0.09}$
& \cellcolor{resultred}$52.59_{0.21}$
& \cellcolor{resultred}$61.25_{0.21}$
& \cellcolor{resultred}$68.28_{0.14}$
& \cellcolor{resultred}$52.77_{0.01}$ \\

& Random Perturb.
& \cellcolor{resultred}$33.39_{0.05}$
& \cellcolor{resultred}$63.44_{0.02}$
& \cellcolor{resultred}$74.86_{0.05}$
& \cellcolor{resultred}$42.16_{0.02}$
& \cellcolor{resultred}$35.02_{0.07}$
& \cellcolor{resultred}$25.33_{0.23}$
& \cellcolor{resultred}$68.59_{0.08}$
& \cellcolor{resultred}$54.15_{0.00}$
& \cellcolor{resultred}$60.83_{0.20}$
& \cellcolor{resultred}$66.49_{0.10}$
& \cellcolor{resultred}$52.43_{0.04}$ \\

& Partial Init.
& \cellcolor{resultred}$34.30_{0.17}$
& \cellcolor{resultred}$64.17_{0.15}$
& \cellcolor{resultred}$70.17_{0.02}$
& \cellcolor{resultred}$42.57_{0.03}$
& \cellcolor{resultred}$33.64_{0.09}$
& \cellcolor{resultred}$24.00_{0.00}$
& \cellcolor{resultred}$68.48_{0.23}$
& \cellcolor{resultred}$58.60_{0.21}$
& \cellcolor{resultred}$62.46_{0.09}$
& \cellcolor{resultred}$66.98_{0.18}$
& \cellcolor{resultred}$52.54_{0.03}$ \\

& \textbf{DS-Upcycling (Ours)}
& \cellcolor{resultgreen}$52.28_{0.10}$
& \cellcolor{resultgreen}$\mathbf{83.01_{0.11}}$
& \cellcolor{resultgreen}$\mathbf{83.30_{0.03}}^*$
& \cellcolor{resultred}$\underline{59.92_{0.01}}$
& \cellcolor{resultgreen}$69.19_{0.03}$
& \cellcolor{resultgreen}$34.40_{0.20}$
& \cellcolor{resultred}$\underline{79.52_{0.13}}$
& \cellcolor{resultgreen}$\mathbf{73.53_{0.42}}^*$
& \cellcolor{resultgreen}$\mathbf{74.40_{0.12}}$
& \cellcolor{resultgreen}$\underline{78.67_{0.20}}$
& \cellcolor{resultgreen}$\mathbf{68.82_{0.04}}^*$ \\

\midrule
\midrule

\multicolumn{1}{c}{}
& \textit{Qwen-3-8B}
& $52.70_{0.25}$
& $81.82_{0.00}$
& $83.08_{0.07}$
& $\underline{58.85_{0.01}}$
& $66.09_{0.07}$
& $32.40_{0.53}$
& $79.18_{0.14}$
& $\mathbf{74.97_{0.21}}$
& $72.40_{0.12}$
& $78.40_{0.10}$
& $67.99_{0.03}$ \\
\midrule

\multirow{4}{*}{\rotatebox[origin=c]{90}{\textbf{Dense}}}
& Simple FT
& \cellcolor{resultgreen}$55.03_{0.23}$
& \cellcolor{resultgreen}$82.25_{0.06}$
& \cellcolor{resultgreen}$84.13_{0.16}$
& \cellcolor{resultred}$53.84_{0.06}$
& \cellcolor{resultred}$59.16_{0.04}$
& \cellcolor{resultred}$29.73_{0.12}$
& \cellcolor{resultred}$76.68_{0.03}$
& \cellcolor{resultred}$66.31_{0.21}$
& \cellcolor{resultred}$68.85_{0.33}$
& \cellcolor{resultred}$73.75_{0.21}$
& \cellcolor{resultred}$64.97_{0.06}$ \\

& LoRA
& \cellcolor{resultgreen}$\mathbf{63.96_{0.34}}$
& \cellcolor{resultgreen}$\mathbf{88.12_{0.11}}$
& \cellcolor{resultgreen}$\underline{86.77_{0.05}}$
& \cellcolor{resultred}$57.87_{0.02}$
& \cellcolor{resultgreen}$\mathbf{68.62_{0.07}}$
& \cellcolor{resultgreen}$32.93_{0.42}$
& \cellcolor{resultgreen}$\underline{79.63_{0.13}}$
& \cellcolor{resultred}$67.75_{0.21}$
& \cellcolor{resultred}$71.61_{0.25}$
& \cellcolor{resultgreen}$\underline{79.09_{0.07}}$
& \cellcolor{resultgreen}$\underline{69.63_{0.06}}$ \\

& LoRAMoE
& \cellcolor{resultgreen}$\underline{62.68_{0.05}}$
& \cellcolor{resultgreen}$\underline{87.35_{0.09}}$
& \cellcolor{resultgreen}$86.48_{0.08}$
& \cellcolor{resultred}$58.45_{0.08}$
& \cellcolor{resultgreen}$\underline{67.49_{0.12}}$
& \cellcolor{resultgreen}$\mathbf{33.87_{0.12}}$
& \cellcolor{resultgreen}$\mathbf{79.87_{0.11}}$
& \cellcolor{resultred}$64.98_{0.00}$
& \cellcolor{resultgreen}$\mathbf{73.90_{0.51}}$
& \cellcolor{resultred}$77.92_{0.25}$
& \cellcolor{resultgreen}$69.30_{0.10}$ \\

& LLaMA-Pro
& \cellcolor{resultgreen}$59.95_{0.26}$
& \cellcolor{resultgreen}$85.54_{0.05}$
& \cellcolor{resultgreen}$85.33_{0.14}$
& \cellcolor{resultred}$56.18_{0.06}$
& \cellcolor{resultred}$58.48_{0.03}$
& \cellcolor{resultgreen}$32.87_{0.42}$
& \cellcolor{resultred}$77.71_{0.03}$
& \cellcolor{resultred}$\underline{73.16_{0.21}}$
& \cellcolor{resultgreen}$73.22_{0.33}$
& \cellcolor{resultred}$76.09_{0.27}$
& \cellcolor{resultred}$67.85_{0.08}$ \\
\midrule

\multirow{5}{*}{\rotatebox[origin=c]{90}{\textbf{Upcycling}}}
& Random Init.
& \cellcolor{resultgreen}$56.48_{0.23}$
& \cellcolor{resultgreen}$83.66_{0.10}$
& \cellcolor{resultgreen}$86.27_{0.03}$
& \cellcolor{resultred}$53.95_{0.03}$
& \cellcolor{resultred}$58.79_{0.08}$
& \cellcolor{resultgreen}$33.47_{0.31}$
& \cellcolor{resultred}$76.84_{0.28}$
& \cellcolor{resultred}$69.55_{0.21}$
& \cellcolor{resultred}$69.69_{0.28}$
& \cellcolor{resultred}$73.53_{0.11}$
& \cellcolor{resultred}$66.22_{0.06}$ \\

& Duplication
& \cellcolor{resultgreen}$54.41_{0.05}$
& \cellcolor{resultgreen}$82.70_{0.04}$
& \cellcolor{resultgreen}$85.12_{0.07}$
& \cellcolor{resultred}$54.11_{0.05}$
& \cellcolor{resultred}$59.56_{0.10}$
& \cellcolor{resultred}$30.60_{0.00}$
& \cellcolor{resultred}$76.41_{0.06}$
& \cellcolor{resultred}$70.40_{0.00}$
& \cellcolor{resultred}$71.40_{0.12}$
& \cellcolor{resultred}$74.23_{0.10}$
& \cellcolor{resultred}$65.89_{0.02}$ \\

& Random Perturb.
& \cellcolor{resultgreen}$56.14_{0.37}$
& \cellcolor{resultgreen}$83.32_{0.12}$
& \cellcolor{resultgreen}$85.73_{0.12}$
& \cellcolor{resultred}$53.83_{0.02}$
& \cellcolor{resultred}$56.21_{0.14}$
& \cellcolor{resultred}$30.80_{0.20}$
& \cellcolor{resultred}$75.99_{0.14}$
& \cellcolor{resultred}$68.95_{0.36}$
& \cellcolor{resultred}$69.59_{0.09}$
& \cellcolor{resultred}$73.97_{0.23}$
& \cellcolor{resultred}$65.45_{0.09}$ \\

& Partial Init.
& \cellcolor{resultgreen}$56.68_{0.13}$
& \cellcolor{resultgreen}$83.42_{0.04}$
& \cellcolor{resultgreen}$86.00_{0.02}$
& \cellcolor{resultred}$53.45_{0.05}$
& \cellcolor{resultred}$55.43_{0.02}$
& \cellcolor{resultred}$31.80_{0.20}$
& \cellcolor{resultred}$76.55_{0.05}$
& \cellcolor{resultred}$69.55_{0.21}$
& \cellcolor{resultred}$69.14_{0.21}$
& \cellcolor{resultred}$73.20_{0.18}$
& \cellcolor{resultred}$65.52_{0.03}$ \\

& \textbf{DS-Upcycling (Ours)}
& \cellcolor{resultgreen}$62.37_{0.23}$
& \cellcolor{resultgreen}$87.09_{0.09}$
& \cellcolor{resultgreen}$\mathbf{87.06_{0.03}}^*$
& \cellcolor{resultgreen}$\mathbf{58.92_{0.05}}$
& \cellcolor{resultgreen}$66.63_{0.07}$
& \cellcolor{resultgreen}$\underline{33.73_{0.31}}$
& \cellcolor{resultgreen}$79.62_{0.19}$
& \cellcolor{resultred}$72.80_{0.21}$
& \cellcolor{resultgreen}$\underline{73.32_{0.08}}$
& \cellcolor{resultgreen}$\mathbf{79.26_{0.14}}^*$
& \cellcolor{resultgreen}$\mathbf{70.08_{0.10}}^*$ \\

\bottomrule
\end{tabular}
}
\caption{Performance on general domain. \textbf{Bold}/\underline{underlined} denote best/second-best results per task. \colorbox[HTML]{D9EAD3}{Green}/\colorbox[HTML]{F4CCCC}{Red} indicates scores above/below the untrained base model. When our method ranks first, $^*$ marks significant improvements over the second-best method (paired $t$-test, $p<0.05$).}
\label{tab:8B-General}
\end{table*}

\subsection{Domain-specific Performance}
\label{sec:ds-performance}

Table~\ref{tab:8B-Specialized} presents the results in the biomedical and legal domains. 

\paragraph{Dense Baselines.}
LoRAMoE performs best among these approaches and improves average performance in both domains.
However, these improvements are accompanied by declines on MedQA (Llama: $60.36 \rightarrow 57.27$; Qwen: $65.54 \rightarrow 62.32$) and Law-MMLU (Llama: $67.98 \rightarrow 67.49$; Qwen: $74.77 \rightarrow 71.79$).
Its domain-level gains do not consistently extend to individual tasks.
One possible explanation is interference among tasks within each domain.
Prior work has found that adapting low-rank adapters to multiple tasks can improve performance on some tasks while degrading others \citep{10.1609/aaai.v39i20.35509}.
DS-Upcycling instead allows full MLP experts to learn domain-specific updates without low-rank constraints.
This greater capacity may help accommodate the different requirements of tasks within each domain.
Our method achieves the highest biomedical (Llama: 68.42; Qwen: 73.34) and legal (Llama: 68.96; Qwen: 72.15) averages across all methods, significantly outperforming the respective second-best baselines (paired $t$-test, $p<0.05$).
These domain-level gains are accompanied by improvements over the corresponding base model on every specialized-domain task.

\paragraph{Upcycling Baselines.}
Upcycling baselines fail to consistently improve performance across specialized domains.
For Llama-3.1-8B, even their highest biomedical (Duplication: $66.18 \rightarrow 50.97$) and legal (Random Init.: $60.30 \rightarrow 56.62$) averages fall below those of the base model.
For Qwen-3-8B, all upcycling baselines improve the legal average but reduce the biomedical average.
These results show that expert initialization alone may not ensure the reliable expert selection and specialization needed for multi-domain adaptation.
For expert selection, prior multilingual upcycling studies have found that explicit routing guidance improves adaptation performance \citep{10.1609/aaai.v39i24.34805,zhang-etal-2025-less}.
For expert specialization, prior MoE research has shown that concentrating task-specific updates within relevant experts supports effective adaptation \citep{wang-etal-2024-expert,li-etal-2025-dynamic}.
DS-Upcycling combines domain-aware router supervision to guide expert selection with domain-specific expert updates to concentrate each domain's learning within its corresponding expert.
We further examine their contributions through routing analysis (\S\ref{sec:routing quality}), expert specialization analysis (\S\ref{sec:Expert Specialization}), and ablations of both components (\S\ref{sec:Ablations}).

\subsection{General Performance}
\label{sec:g-performance}

Table~\ref{tab:8B-General} presents the results on general-domain tasks. Overall, existing baselines struggle to consistently preserve general-domain capabilities.

\paragraph{Dense Baselines.}
LoRAMoE is the only dense method that improves average performance in both model families (Llama: 67.92; Qwen: 69.30).
However, it still degrades five out of ten tasks on Llama-3.1-8B and three out of ten on Qwen-3-8B.
RTE performance declines substantially in both cases (Llama: $70.52 \rightarrow 65.82$; Qwen: $74.97 \rightarrow 64.98$).
One possible explanation is that freezing pre-trained parameters does not prevent domain-specific adapter updates from affecting general-domain performance.
Although LoRAMoE freezes pre-trained weights, it still combines all adapter outputs for general-domain inputs.
DS-Upcycling uses domain-aware router supervision to favor the general expert for these inputs, thereby reducing interference from specialized domains.
Our ablation study supports this interpretation, showing that removing domain-aware router supervision reduces general-domain performance on both model families (\S\ref{sec:Ablations}).
DS-Upcycling achieves the highest general-domain averages (Llama: 68.82; Qwen: 70.08) across all methods, significantly outperforming the respective second-best baselines (paired $t$-test, $p<0.05$).
Only two tasks decline slightly on Llama-3.1-8B (HellaSwag: $60.02 \rightarrow 59.92$; PIQA: $80.09 \rightarrow 79.52$).
For Qwen-3-8B, only RTE declines ($74.97 \rightarrow 72.80$).
These results indicate that DS-Upcycling preserves general-domain capabilities more effectively.

\paragraph{Upcycling Baselines.}
All upcycling baselines reduce average general-domain performance on both model families.
Even their highest averages fall below those of the base models (Llama-3.1-8B, Duplication: $67.79 \rightarrow 52.77$; Qwen-3-8B, Random Init.: $67.99 \rightarrow 66.22$).
Updating all parameters during specialized-domain adaptation may alter pre-trained knowledge.
Duplication falls to 52.77 for Llama and 65.89 for Qwen despite initially matching the dense model's outputs.
Preserving initial outputs alone might be insufficient to retain general-domain capabilities after adaptation.
DS-Upcycling instead freezes pre-trained parameters and updates only the corresponding expert and router during adaptation to each domain.
Our ablation further supports the contribution of domain-specific expert updates to retaining general-domain performance (\S\ref{sec:Ablations}).
Even with the general expert frozen, updating all specialized experts lowers general-domain averages on both model families.
In general, we observe that freezing pre-trained parameters and restricting expert updates to their corresponding domains help mitigate general-domain catastrophic forgetting.

\section{Analysis}

\subsection{Routing Quality}
\label{sec:routing quality}

To evaluate whether domain-aware router supervision (\S\ref{sec:step-2}) helps routers distinguish input domains, we compare DS-Upcycling with upcycling baselines that randomly initialize their routers. We sample 128 additional examples per domain (general, legal and biomedical) with no overlap with those used for router supervision. We measure routing accuracy in the upcycled models before and after adaptation by checking if the expert corresponding to the input domain is among the selected experts.

Table~\ref{tab:router acc} reports routing accuracy across domains. For Duplication, routing accuracy remains largely unchanged across all three domains after adaptation (Llama general: $63.75 \rightarrow 62.62$, legal: $69.31 \rightarrow 69.56$, and biomedical: $63.00 \rightarrow 63.56$). The other upcycling baselines become strongly biased toward the general domain while routing accuracy on legal and biomedical inputs declines substantially (Random Init.\ on Llama general: $68.31 \rightarrow 99.62$, legal: $62.94 \rightarrow 51.00$, and biomedical: $61.06 \rightarrow 42.69$). This pattern shows that training on domain-specific data alone does not enable randomly initialized routers to reliably select the corresponding specialized experts. This routing failure contributes to the specialized-domain performance degradation observed in \S\ref{sec:ds-performance}.

\begin{table}[ht]
\centering
\renewcommand{\arraystretch}{1}
\footnotesize
\setlength{\tabcolsep}{1mm}
\resizebox{\columnwidth}{!}{
\begin{tabular}{clcccccc}
\toprule
\multicolumn{2}{c}{}
& \multicolumn{2}{c}{\textbf{General}}
& \multicolumn{2}{c}{\textbf{Legal}}
& \multicolumn{2}{c}{\textbf{Biomedical}} \\
\cmidrule(l{2pt}r{2pt}){3-4}
\cmidrule(l{2pt}r{2pt}){5-6}
\cmidrule(l{2pt}r{2pt}){7-8}
& \multicolumn{1}{c}{\textbf{Methods}}
& Before & After
& Before & After
& Before & After \\
\midrule

\multicolumn{1}{c|}{
\multirow{5}{*}{\rotatebox[origin=c]{90}{\textbf{Llama}}}}
& Random Init.
& 68.31 & 99.62 & 62.94 & 51.00 & 61.06 & 42.69 \\

\multicolumn{1}{c|}{}
& Duplication
& 63.75 & 62.62 & 69.31 & 69.56 & 63.00 & 63.56 \\

\multicolumn{1}{c|}{}
& Random Perturb.
& 63.56 & 99.94 & 69.44 & 48.25 & 73.88 & 57.56 \\

\multicolumn{1}{c|}{}
& Partial Init.
& 63.75 & 99.81 & 69.31 & 53.75 & 63.00 & 45.62 \\

\multicolumn{1}{c|}{}
& \textbf{DS-Upcycling (Ours)}
& 100.00 & 100.00 & 99.06 & 98.12 & 97.62 & 95.37 \\

\midrule
\midrule

\multicolumn{1}{c|}{
\multirow{5}{*}{\rotatebox[origin=c]{90}{\textbf{Qwen}}}}
& Random Init.
& 55.28 & 99.33 & 75.83 & 46.50 & 72.39 & 54.56 \\

\multicolumn{1}{c|}{}
& Duplication
& 69.89 & 68.94 & 73.44 & 69.89 & 62.06 & 64.44 \\

\multicolumn{1}{c|}{}
& Random Perturb.
& 65.72 & 99.61 & 69.78 & 49.83 & 74.28 & 55.22 \\

\multicolumn{1}{c|}{}
& Partial Init.
& 64.61 & 99.72 & 70.61 & 45.39 & 71.00 & 55.22 \\

\multicolumn{1}{c|}{}
& \textbf{DS-Upcycling (Ours)}
& 100.00 & 100.00 & 89.78 & 91.44 & 97.72 & 89.94 \\

\bottomrule
\end{tabular}
}
\caption{Routing accuracy before and after adaptation across general, legal, and biomedical domains.}
\label{tab:router acc}
\end{table}

In contrast, DS-Upcycling achieves high routing accuracy across all three domains and maintains it after adaptation. This routing stability indicates that domain-aware router supervision enables consistent selection of appropriate experts across domains. This explains the consistent specialized-domain performance gains achieved by our method.

\subsection{Expert Specialization}
\label{sec:Expert Specialization}

We next evaluate whether restricting each expert's updates to its corresponding domain (\S\ref{sec:step-4}) promotes specialization despite identical initialization. Following prior work \citep{li2017understandingneuralnetworksrepresentation}, we compute the cosine similarity between the biomedical and legal experts' parameter updates as: $\mathrm{Cos}\left(W_{\mathrm{Bio}}^{\mathrm{trained}} - W_{\mathrm{Bio}}^{\mathrm{init}}, W_{\mathrm{Legal}}^{\mathrm{trained}} - W_{\mathrm{Legal}}^{\mathrm{init}}\right)$.

Table~\ref{Tab: Update-Similarity} shows the results. Lower similarity indicates less aligned update directions and serves as an indicator of expert specialization. Duplication yields high update similarity across both model families. This suggests that its experts have similar parameter updates despite being intended for different domains. Other upcycling baselines achieve lower similarity but alter the pre-trained model's outputs at initialization. This can hinder subsequent adaptation \citep{wu-etal-2024-llama, cao2025progressivedepthupscalingoptimal} and contribute to the general-domain degradation observed in \S\ref{sec:g-performance}. In contrast, DS-Upcycling starts from the same copied weights as Duplication yet achieves low update similarity comparable to the other upcycling baselines. Restricting each expert's updates to its corresponding domain therefore enables identically initialized experts to specialize in distinct domains.

\begin{table}[ht]
\centering
\renewcommand{\arraystretch}{1.15}
\footnotesize
\setlength{\tabcolsep}{1mm}
\resizebox{0.6\columnwidth}{!}{
\begin{tabular}{c|lc}
\toprule
\multicolumn{1}{c}{} & \multicolumn{1}{c}{\textbf{Methods}}
& \textbf{Similarity $\downarrow$} \\
\midrule

\multirow{5}{*}{\rotatebox[origin=c]{90}{\textbf{Llama}}}
& Random Init.        & 0.03\% \\
& Duplication         & 36.83\% \\
& Random Perturb.     & 0.21\% \\
& Partial Init.       & 0.38\% \\
& \textbf{DS-Upcycling (Ours)}       & 0.59\% \\

\midrule
\midrule

\multirow{5}{*}{\rotatebox[origin=c]{90}{\textbf{Qwen}}}
& Random Init.        & 0.01\% \\
& Duplication         & 31.06\% \\
& Random Perturb.     & 0.46\% \\
& Partial Init.       & 1.07\% \\
& \textbf{DS-Upcycling (Ours)}       & 0.76\% \\

\bottomrule
\end{tabular}
}
\caption{Cosine similarity between biomedical and legal expert parameter updates after adaptation.}
\label{Tab: Update-Similarity}
\end{table}

\subsection{Ablation Study}
\label{sec:Ablations}

\paragraph{Domain-Aware Router Supervision (\S\ref{sec:step-2}).} We compare DS-Upcycling with a variant that randomly initializes routers without domain-aware supervision while keeping all other components unchanged. Table~\ref{tab:ablation} shows that domain-aware router supervision improves average performance in five of the six settings spanning two model families and three domains. The only exception is the biomedical domain on Qwen-3-8B where the average decreases slightly from 73.46 to 73.34. These results validate the contribution of domain-aware router supervision to effective multi-domain adaptation.

\begin{table}[!ht]
\centering
\renewcommand{\arraystretch}{1.2}
\setlength{\tabcolsep}{1mm}
\resizebox{\linewidth}{!}{
\begin{tabular}{l|lccc}
\toprule
\multicolumn{1}{c}{} & \multicolumn{1}{c}{\textbf{Methods}}
& \textbf{Bio} & \textbf{Leg} & \textbf{Gen} \\
\midrule

\multirow{3}{*}{\rotatebox[origin=c]{90}{\textbf{Llama}}}
& DS-Upcycling
& $\mathbf{68.42_{0.08}}^*$
& $\mathbf{68.96_{0.14}}^*$
& $\mathbf{68.82_{0.04}}^*$ \\
& ~w/o DA Router Superv.
& $\underline{67.15_{0.17}}$
& $66.57_{0.10}$
& $\underline{67.87_{0.03}}$ \\
& ~w/o DS Expert Upd.
& $62.31_{0.02}$
& $\underline{67.98_{0.18}}$
& $64.33_{0.03}$ \\

\midrule
\midrule

\multirow{3}{*}{\rotatebox[origin=c]{90}{\textbf{Qwen}}}
& DS-Upcycling
& $\underline{73.34_{0.09}}$
& $\mathbf{72.15_{0.09}}^*$
& $\mathbf{70.08_{0.10}}^*$ \\
& ~w/o DA Router Superv.
& $\mathbf{73.46_{0.09}}$
& $69.83_{0.08}$
& $\underline{68.89_{0.03}}$ \\
& ~w/o DS Expert Upd.
& $70.96_{0.06}$
& $\underline{71.25_{0.06}}$
& $68.63_{0.03}$ \\

\bottomrule
\end{tabular}
}
\caption{Ablation results. ``w/o DA Router Superv.'' and ``w/o DS Expert Upd.'' evaluate the contributions of domain-aware router supervision (\S~\ref{sec:step-2}) and domain-specific expert updates (\S~\ref{sec:step-4}), respectively. Detailed results are presented in Tables~\ref{tab:ablation-specialized} and~\ref{tab:ablation-general} in Appendix~\ref{app:ablation}.}
\label{tab:ablation}
\end{table}

\paragraph{Domain-Specific Expert Updates (\S\ref{sec:step-4}).} We evaluate our strategy of restricting each expert's updates to its corresponding domain by comparing DS-Upcycling with a variant that updates all specialized experts during adaptation to each domain. Table~\ref{tab:ablation} shows that restricting updates to the corresponding expert improves average performance across all six settings. These results validate the effectiveness of promoting expert specialization through domain-specific expert updates.

\section{Conclusion}

In this work, we proposed DS-Upcycling for adapting pre-trained language models to multiple specialized domains. It adds full MLP experts to learn domain-specific knowledge without low-rank constraints and a router to selectively activate relevant experts. Domain-aware router supervision encourages the selection of the general expert for general-domain inputs and suppresses irrelevant experts for specialized-domain inputs. Domain-specific expert updates promote specialization by updating only the corresponding expert and router during adaptation to each domain while freezing all other parameters. Our experiments show that DS-Upcycling offers the highest average performance across specialized and general domains. After adaptation, it achieves 90--98\% routing accuracy on specialized domains compared with 43--70\% for existing upcycling baselines. Our analysis suggest that domain-aware router supervision and domain-specific expert updates help prioritize relevant domain knowledge while minimizing interference from irrelevant domains. These findings highlight the potential of upcycling for effective multi-domain adaptation.

\section*{Limitations}

DS-Upcycling demonstrates strong multi-domain adaptation performance and mitigates forgetting. Unlike low-rank adaptation methods, \textsc{DS-Upcycling} allocates full MLP experts to prevent the expressivity bottlenecks common in multi-task LoRA. Consequently, while inference computation and latency remain bounded by sparse Top-2 activation, the total static parameter footprint stored in GPU memory scales linearly with the number of instantiated domains. An intersting avenue for future work is to integrate our approach with post-adaptation expert quantization or pruning. This would optimize the storage footprint in memory-constrained serving environments supporting multiple domains.

\bibliography{custom}

\appendix

\section{Implementation details}
\label{App:Imp detail}

Following \citet{switch_transformers} and \citet{cao2026fundamentalreasoningparadigmsinduce}, we upcycle the base model's odd-numbered layers so that half become sparse MoE layers. Each upcycled layer retains the original MLP as the general expert and adds two specialized experts for the legal and biomedical domains. We use Top-2 routing to activate the two experts with the highest routing probabilities \citep{switch_transformers}.

For \textit{Random Initialization}, we initialize new expert weights from $\mathcal{N}(0,0.02^2)$ following \citet{yang2025qwen3technicalreport}. For \textit{Random Perturbation}, we replace 50\% of each new expert's weights with values sampled from $\mathcal{N}(0,0.02^2)$ following \citet{muennighoff2025olmoe}. For \textit{Partial Initialization}, we re-initialize 50\% of each new expert's weights using Gaussian distributions matched to their original mean and variance following \citet{ICLR2025_d24b7366}.

We use similar parameter counts across the multi-domain adaptation baselines for a fair comparison. All these baselines update only newly added parameters except Simple FT. Each LoRA adapter has rank 1024 for LoRA and 1536 for LoRAMoE. For LLaMA-Pro, we divide the base model into four layer groups and insert seven new transformer layers per group.

\section{Hyper-parameter Details}
\label{App:hyp detail}

We use a global batch size of 64 and a sequence length of 4096. For both domains, we use a cosine learning rate schedule with warm-up and decay from $5\times10^{-5}$ to $5\times10^{-6}$. During biomedical adaptation, we lower this range to $5\times10^{-6}$--$5\times10^{-7}$ for the upcycling baselines and Simple FT because the original range yielded poor performance in our preliminary experiments. We fine-tune all models on a single NVIDIA GH200 (96GB) using Flash-Attention 2 \citep{dao2023flashattention2} and \texttt{bf16} mixed precision. We evaluate all models zero-shot using the Language Model Evaluation Harness \citep{eval-harness}.

\section{Ablation Results}
\label{app:ablation}

Tables~\ref{tab:ablation-specialized} and~\ref{tab:ablation-general} present the full ablation results on specialized and general domains, respectively.

\setlength{\intextsep}{\originalintextsep}
\begin{table*}[!ht]
\begin{center}
\renewcommand{\arraystretch}{1.8}
\footnotesize
\setlength{\tabcolsep}{1mm}
\resizebox{0.95\textwidth}{!}{
\begin{tabular}{c|lccccccccc}
\toprule
\multicolumn{2}{c}{} & \multicolumn{5}{c}{\textbf{Biomedical}}
& \multicolumn{4}{c}{\textbf{Legal}} \\
\cmidrule(l{2pt}r{2pt}){3-7} \cmidrule(l{2pt}r{2pt}){8-11}
\multicolumn{1}{c}{} & \multicolumn{1}{c}{\textbf{Methods}}
& \textbf{MedMCQA} & \textbf{MedQA} & \textbf{Bio-MMLU} & \textbf{PubMedQA} & \textbf{Avg}
& \textbf{Law-MMLU} & \textbf{CaseHOLD} & \textbf{ECtHR} & \textbf{Avg} \\
\midrule

\multirow{3}{*}{\rotatebox[origin=c]{90}{\textbf{Llama-3.1-8B}}}
& DS-Upcycling
& $\mathbf{59.76_{0.11}}^*$
& $\mathbf{61.51_{0.34}}^*$
& $\mathbf{75.08_{0.05}}^*$
& $\mathbf{77.33_{0.12}}$
& $\mathbf{68.42_{0.08}}^*$
& $\mathbf{68.79_{0.31}}$
& $\underline{75.52_{0.10}}$
& $\underline{62.58_{0.26}}$
& $\mathbf{68.96_{0.14}}^*$ \\
& ~w/o DA Router Superv.
& $\underline{58.92_{0.18}}$
& $\underline{59.91_{0.12}}$
& $\underline{72.64_{0.14}}$
& $\underline{77.13_{0.31}}$
& $\underline{67.15_{0.17}}$
& $\underline{68.62_{0.37}}$
& $72.30_{0.14}$
& $58.79_{0.12}$
& $66.57_{0.10}$ \\
& ~w/o DS Expert Upd.
& $53.83_{0.12}$
& $52.00_{0.21}$
& $66.86_{0.16}$
& $76.53_{0.42}$
& $62.31_{0.02}$
& $57.39_{0.47}$
& $\mathbf{83.23_{0.07}}$
& $\mathbf{63.33_{0.14}}$
& $\underline{67.98_{0.18}}$ \\

\midrule
\midrule

\multirow{3}{*}{\rotatebox[origin=c]{90}{\textbf{Qwen-3-8B}}}
& DS-Upcycling
& $\underline{66.62_{0.05}}$
& $\mathbf{69.05_{0.27}}$
& $\underline{80.41_{0.22}}$
& $\underline{77.27_{0.31}}$
& $\underline{73.34_{0.09}}$
& $\mathbf{75.76_{0.16}}^*$
& $\underline{78.12_{0.11}}$
& $\underline{62.58_{0.14}}$
& $\mathbf{72.15_{0.09}}^*$ \\
& ~w/o DA Router Superv.
& $\mathbf{67.18_{0.22}}$
& $\underline{68.53_{0.23}}$
& $\mathbf{81.13_{0.17}}$
& $77.00_{0.20}$
& $\mathbf{73.46_{0.09}}$
& $\underline{75.12_{0.10}}$
& $71.93_{0.15}$
& $62.46_{0.19}$
& $69.83_{0.08}$ \\
& ~w/o DS Expert Upd.
& $60.89_{0.09}$
& $65.72_{0.05}$
& $79.23_{0.09}$
& $\mathbf{78.00_{0.20}}$
& $70.96_{0.06}$
& $69.70_{0.08}$
& $\mathbf{81.21_{0.04}}$
& $\mathbf{62.83_{0.19}}$
& $\underline{71.25_{0.06}}$ \\

\bottomrule
\end{tabular}
}
\end{center}
\caption{Ablation performance on specialized domains.}
\label{tab:ablation-specialized}
\end{table*}

\begin{table*}[!ht]
\begin{center}
\renewcommand{\arraystretch}{1.8}
\footnotesize
\setlength{\tabcolsep}{1mm}
\resizebox{0.95\textwidth}{!}{
\begin{tabular}{c|lccccccccccc}
\toprule
\multicolumn{1}{c}{} & \multicolumn{1}{c}{\textbf{Methods}}
& \textbf{ARC-C} & \textbf{ARC-E} & \textbf{BoolQ}
& \textbf{HellaSwag} & \textbf{LAMBADA} & \textbf{OpenBookQA}
& \textbf{PIQA} & \textbf{RTE} & \textbf{WinoGrande}
& \textbf{StoryCloze} & \textbf{Avg} \\
\midrule

\multirow{3}{*}{\rotatebox[origin=c]{90}{\textbf{Llama-3.1-8B}}}
& DS-Upcycling
& $\underline{52.28_{0.10}}$
& $\underline{83.01_{0.11}}$
& $\mathbf{83.30_{0.03}}$
& $\mathbf{59.92_{0.01}}^*$
& $\mathbf{69.19_{0.03}}^*$
& $\underline{34.40_{0.20}}$
& $\mathbf{79.52_{0.13}}^*$
& $\mathbf{73.53_{0.42}}^*$
& $\underline{74.40_{0.12}}$
& $\underline{78.67_{0.20}}$
& $\mathbf{68.82_{0.04}}^*$ \\
& ~w/o DA Router Superv.
& $\mathbf{53.75_{0.09}}$
& $\mathbf{83.49_{0.10}}$
& $\underline{83.23_{0.21}}$
& $\underline{59.38_{0.04}}$
& $\underline{68.22_{0.09}}$
& $\mathbf{36.80_{0.20}}$
& $\underline{79.05_{0.09}}$
& $60.65_{0.36}$
& $\mathbf{74.93_{0.20}}$
& $\mathbf{79.22_{0.18}}$
& $\underline{67.87_{0.03}}$ \\
& ~w/o DS Expert Upd.
& $46.13_{0.13}$
& $76.67_{0.13}$
& $79.57_{0.11}$
& $56.00_{0.03}$
& $61.46_{0.09}$
& $32.27_{0.12}$
& $76.97_{0.03}$
& $\underline{69.55_{0.42}}$
& $70.35_{0.16}$
& $74.37_{0.04}$
& $64.33_{0.03}$ \\

\midrule
\midrule

\multirow{3}{*}{\rotatebox[origin=c]{90}{\textbf{Qwen-3-8B}}}
& DS-Upcycling
& $\mathbf{62.37_{0.23}}^*$
& $\mathbf{87.09_{0.09}}^*$
& $\mathbf{87.06_{0.03}}^*$
& $\mathbf{58.92_{0.05}}^*$
& $\underline{66.63_{0.07}}$
& $\mathbf{33.73_{0.31}}^*$
& $\mathbf{79.62_{0.19}}$
& $\underline{72.80_{0.21}}$
& $\mathbf{73.32_{0.08}}^*$
& $\mathbf{79.26_{0.14}}$
& $\mathbf{70.08_{0.10}}^*$ \\
& ~w/o DA Router Superv.
& $58.93_{0.13}$
& $84.89_{0.13}$
& $85.56_{0.06}$
& $\underline{58.82_{0.08}}$
& $\mathbf{66.66_{0.14}}$
& $\underline{32.60_{0.00}}$
& $\underline{79.31_{0.03}}$
& $71.00_{0.42}$
& $\underline{72.45_{0.21}}$
& $\underline{78.67_{0.14}}$
& $\underline{68.89_{0.03}}$ \\
& ~w/o DS Expert Upd.
& $\underline{60.35_{0.13}}$
& $\underline{86.55_{0.05}}$
& $\underline{86.52_{0.08}}$
& $57.34_{0.06}$
& $64.02_{0.05}$
& $31.07_{0.31}$
& $78.75_{0.23}$
& $\mathbf{73.04_{0.42}}$
& $71.69_{0.05}$
& $76.99_{0.17}$
& $68.63_{0.03}$ \\

\bottomrule
\end{tabular}
}
\end{center}
\caption{Ablation performance on general domain.}
\label{tab:ablation-general}
\end{table*}

\section{Use of Generative AI Tools}
The authors acknowledge the use of GPT-6 Astra during the preparation of this manuscript. Its use is limited to improving the grammar and clarity of the original draft.

\end{document}